\documentclass[10pt,twocolumn,letterpaper]{article}

\usepackage{iccv}
\usepackage{times}
\usepackage{epsfig}
\usepackage{graphicx}
\usepackage{amsmath}
\usepackage{amssymb}

\usepackage[breaklinks=true,bookmarks=false]{hyperref}

\iccvfinalcopy 

\def\iccvPaperID{****} 

\usepackage{subfigure}
\usepackage[font=small,labelfont=bf]{caption}
\graphicspath{{figures/}}

\begin{document}

\makeatletter
\let\@oldmaketitle\@maketitle
\renewcommand{\@maketitle}{\@oldmaketitle
    \vspace{-0.5cm}

\includegraphics[width=\linewidth]{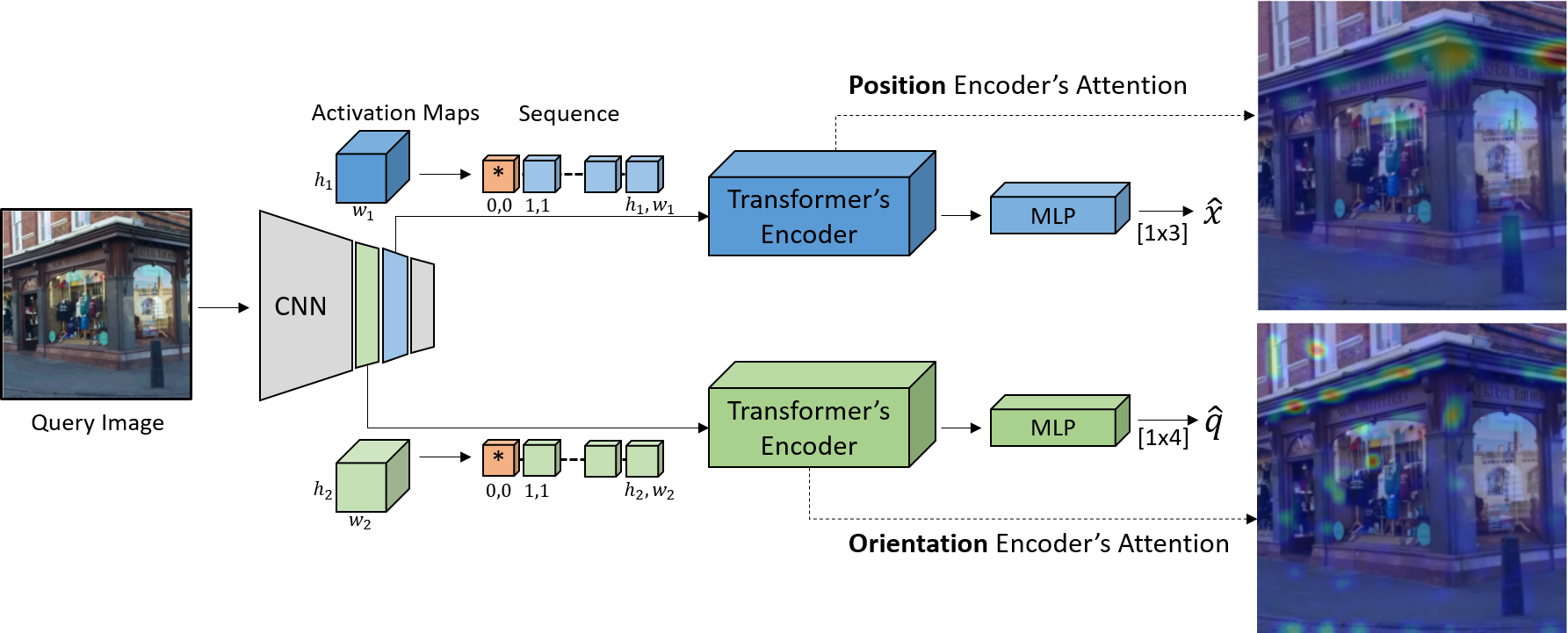}
    \captionof{figure}{The proposed attention-based regression localization scheme. The input image is
		first encoded by a convolutional backbone. Two activation maps, at different resolutions, are transformed into sequential representations. The two activation sequences are analyzed by dual Transformer encoders, one per regression task. We depict the attention weights via
		heatmaps. Position is best estimated by corner-like image features,
		while orientation is estimated by edge-like features. Each Transformer encoder output is  used to regress the respective camera pose component (position $\hat{x}$ or orientation $\hat{q}$).}
    \label{fig:teaser}
    \bigskip
    }
\makeatother

\title{Paying Attention to Activation Maps in Camera Pose Regression}
\author{Yoli Shavit\thanks{Denotes equal contribution}
\and Ron Ferens\footnotemark[1]\\
Bar-Ilan University, Israel
\and Yosi Keller
}
\maketitle

\begin{abstract}
Camera pose regression methods apply a single forward pass to the query
image to estimate the camera pose. As such, they offer a fast and
light-weight alternative to traditional localization schemes based on image
retrieval. Pose regression approaches simultaneously learn two regression
tasks, aiming to jointly estimate the camera position and orientation using
a single embedding vector computed by a convolutional backbone. We propose
an attention-based approach for pose regression, where the convolutional
activation maps are used as sequential inputs. Transformers are applied to
encode the sequential activation maps as latent vectors, used for camera
pose regression. This allows us to pay attention to spatially-varying deep
features. Using two Transformer heads, we separately focus on the features
for camera position and orientation, based on how informative they are per
task. Our proposed approach is shown to compare favorably to contemporary
pose regressors schemes and achieves state-of-the-art accuracy across
multiple outdoor and indoor benchmarks. In particular, to the best of our
knowledge, our approach is the only method to attain sub-meter average
accuracy across outdoor scenes. We make our code publicly available from: \href{https://anonymous.4open.science/r/d2ac8b5e-c2d0-4550-9914-9bb4065a9edb/}{here}.
\end{abstract}



\section{Introduction}

Estimating the camera position and orientation using a captured image is an
essential task in various contemporary computer vision applications, ranging
from augmented reality to autonomous driving to navigation in closed
environments such as malls, museums, airports and amusement parks, to name a
few. Many of these applications require a fast and light-weight solution
that is deployable on low-end devices, such as cellphones.

Recent \textit{state-of-the-art} pipelines for visual pose estimation follow
a hierarchical paradigm \cite%
{sattler2016efficient,taira2018inloc,sarlin2019coarse}, where given a query
image, an image retrieval (IR) method is first applied to fetch candidate
images from a large, geo-referenced image database. Local features are
extracted from the query and geo-referenced images and then matched. The
resulting 2D-2D matches are transformed into 2D-3D correspondences using
depth and/or a 3D model obtained with methods such as Light Detection and
Ranging (LiDAR) and Structure from Motion (SfM). Finally, the pose (or a set
of poses) is estimated from 2D-3D matches using Perspective-n-Point (PnP)
and RANSAC algorithms \cite{fischler1981random}. While recent pipelines
implementing the aforementioned approach achieve remarkable accuracy, they
entail significant difficulties. First, such systems have to be implemented
in a client-server architecture, due to the large image datasets that cannot
be stored on thin clients (cellphone etc.). Second, a client-server approach
requires an online data connection which is inapplicable for standalone
localization applications. Last, the localization typically requires $%
50-2000ms$ processing time, depending on the method used, the size of the
database and the number of matches \cite{sarlin2019coarse}. In contrast,
absolute regression-based approaches estimate the camera pose with a single
forward pass, using the query image as an input to a trained regressor. The
end device has to store only the regressor parameters.

As convolutional neural networks (CNNs) were shown to be efficient and
accurate regressors in multiple domains, they were also applied to camera
pose regression, starting with the seminal work of Kendall et al. \cite%
{kendall2015posenet}. The proposed architecture consisted of a convolutional
backbone and a multilayer perceptron (MLP) head, separately regressing the
position and orientation. The appealing 5 $ms$ runtime and simplicity (a
single component instead of a heavy pipeline) paved the way to a new
research paradigm for camera pose estimation. Numerous methods, soon to
follow, aimed at maintaining the low runtime and memory requirements, while
improving the accuracy and generalization of the original method \cite%
{kendall2015posenet,kendall2016modelling,kendall2017geometric,melekhov2017image,naseer2017deep,walch2017image,wu2017delving,shavitferensirpnet,wang2020atloc,cai2019hybrid}%
.

Different camera pose regressors suggested different backbones \cite%
{melekhov2017image,shavitferensirpnet}, loss formulations \cite%
{kendall2017geometric} and MLP architectures \cite%
{wu2017delving,naseer2017deep} as well as additional manipulations of the
output \cite{walch2017image,wang2020atloc}. Common to all these methods, is
that they apply the pose regression using a \textit{single} global image
encoding computed by the backbone CNN.

In this work, we propose a novel attention-based approach that considers an
activation map as a sequence of local spatial embeddings. Specifically, we employ Transformers \cite{AttentionIsAllYouNeed}, shown to provide an efficient
computational means for analyzing spatially varying image activations as
sequences \cite{DETR}. This allows for an
adaptive analysis of the spatial image content, as images typically contain
a mixture of localization-informative and uninformative image cues.
Thus, we cast the camera pose regression as two sequence-to-one problems,
where the intermediate activation maps are input to a pair of
translation-aware and rotation-aware Transformer encoders (Fig. \ref{fig:teaser}). The output of each
encoder is used to regress the corresponding localization attribute. The gist of our approach is that each
Transformer Encoder captures different informative
image cues for either position or orientation estimation. We demonstrate this by overlaying the attention weights as heatsmaps (Figs. \ref{fig:teaser} and \ref{fig3_attention_maps}). The largest attention values of the Positional Encoder are around corners, which are a commonly used localization cue, as in the SIFT detector \cite{SIFT}. Similarly, the Orientational
Encoder emphasizes elongated edges that are informative visual cues for
orientation estimation. Other image features that are less location-informative show
weak attention activations. By adaptively detecting and weighting the
informative spatial activations, Encoders improve the localization accuracy.

In summary, our contributions are as follows:

\begin{itemize}
\item We propose a novel attention-based approach to camera pose regression
using dual encoders, one per regression task.

\item Our approach is experimentally shown to compare favourably with
\textit{state-of-the-art} regression-based schemes, when applied to outdoor
and indoor localization datasets.
\end{itemize}

\section{Related Work}

\subsection{Image-Based Camera Pose Estimation}

\label{subsec:related pose estimation}

A common approach to image-based pose estimation applies large-scale image
retrieval followed by image matching and PnP-RANSAC \cite%
{taira2018inloc,sarlin2019coarse,noh2017large,dusmanu2019d2}. Each query
image to be localized, is encoded using a CNN trained for IR, and a
relatively small set of nearest neighbors is retrieved for image-to-image
matching. Local feature matching can be applied using brute force methods
\cite{taira2018inloc} or another learned component \cite{sarlin2020superglue}%
. In order to convert image matches to the world plane, spatial information
needs to be available at inference time. The pose can then be estimated
using PnP-RANSAC from 2D-3D correspondences. In this work we study a faster
and lighter alternative to this pipelined approach, namely camera pose
regression.

Camera pose regression was first suggested by \cite{kendall2015posenet}. The
authors proposed to follow the success of CNN backbones in extracting visual
features from images and apply them for regressing the pose directly from
the input image. Specifically, an MLP head was attached to a GoogLeNet
backbone, regressing the position and orientation of the camera (replacing
the classifier heads in the original architecture). This novel approach,
albeit far less accurate than localization pipelines, enabled performing
pose estimation with a single forward pass in only a few milliseconds.

To the best of our knowledge, the paradigm of regressing the camera pose
from the final output of a CNN backbone was adopted by all regressors to
date \cite{shavit2019introduction}. Variations to the architecture focused
on alternatives to the original proposed CNN backbone \cite%
{melekhov2017image,naseer2017deep,wu2017delving, shavitferensirpnet} and on
deeper, branching architectures for the MLP head \cite%
{wu2017delving,naseer2017deep}. Other works tried to address overfitting by
averaging over predictions from models with randomly dropped activations
\cite{kendall2016modelling} or by reducing the dimensionality of the global
image encoding with Long-Short-Term-Memory (LSTM) layers \cite%
{walch2017image}. Multi-modality fusion (for example, with inertial sensors)
was also suggested as a means to improve accuracy \cite%
{brahmbhatt2018geometry}. Recently, Wang et al. suggested to use attention
in order to guide the regression process \cite{wang2020atloc}. The authors
applied dot product self-attention with SoftMax on the final output of the
CNN backbone and updated it with the new attention-based representation
(through summation), before regressing the pose with an MLP head. While our
work also uses attention, we propose a novel paradigm, where instead of
directly regressing the pose from a single global embedding (with/without
attention-guided weights) multi-resolution activations maps are processed as
a sequence of features using Transformer encoders.

In terms of the learning process, one of the main challenges is to
appropriately weight the losses associated with the position and orientation
predictions. Kendall et al. suggested to learn the parameters controlling
the tradeoff between the losses to gain better performance \cite%
{kendall2017geometric}. This formulation was adopted by many pose
regressors, however it still requires manually tuning the parameters'
initialization for different datasets \cite{valada2018deep}. In a recent
work \cite{shavitferensirpnet}, the authors trained the model separately for
position and orientation in order to reduce the need of additional
parameters, while achieving comparable accuracy. Alternative representations
for the orientation were also proposed to gain better balance and stability
of the pose loss \cite{wu2017delving,brahmbhatt2018geometry}. As part of
studying the performance of pose regressors, Sattler et al. proposed an
IR-based baseline for camera pose regression \cite{sattler2019understanding}%
. With this baseline, pose predictions are made by taking a weighted average
of poses from a set of nearest neighbors. The weighting is computed during
inference time based on the distances between the IR encoding of the query
and its neighbors. This baseline was used as an empirical illustration for
some of the limitations of current pose regressors since no regressor was
able to consistently surpass it on either outdoor or indoor datasets (we use
the same datasets for evaluating our work).

Brachmann and Rother introduced a novel class of pose regression schemes
\cite{DSAC,DSAC++}, by training a CNN to estimate the 3D locations
corresponding to the pixels in the query image. This establishes 2D-3D
correspondences that are utilized by a differentiable PnP-RANSAC to estimate
the camera pose, in both training and testing phases. Such approaches
achieve state-of-the-art accuracy. However, since PnP-RANSAC is applied at
inference time, they are significantly slower compared to purely
regression-based approaches. A different, yet related body of works,
estimate the absolute camera pose using the relative motion between the
query image and a reference image, for which the ground truth pose is known.
The learning thus focuses on regressing the relative pose given a pair of
images \cite{balntas2018relocnet,laskar2017camera,ding2019camnet}. These
methods can generalize better since the model is no longer confined to the
absolute reference scene, but require the availability of a pose-labelled
database of anchors during inference time. Combining relative and absolute
regression was also shown to achieve impressive accuracy \cite%
{valada2018deep,radwan2018vlocnet++}.

\subsection{Attention and Transformers in Image Analysis}

Transformers were introduced by Vaswani et al. \cite{AttentionIsAllYouNeed}
as a novel formulation of attention-based sequence encoding and analysis.
Attention mechanisms \cite{DBLP:journals/corr/BahdanauCB14} are neural
network layers that aggregate information from the entire input sequence.
Transformers introduced attention layers, that scan through each element of
a sequence and update it by aggregating information from the whole sequence.
Attention-based approaches were shown to outperform RNNs in encoding long
sequences, and were applied in multiple recent works in natural language
processing (NLP) and computer vision \cite{bert,radford2019language}. In
particular, in the context of image analysis, the use of attention relates
to the bilinear pooling approach of Zhang et al.\cite{bilipool}, and its
multiple extensions \cite{bilipool1}. These schemes compute the inner
products between all CNN activations, and were shown to improve fine-grained
recognition significantly. Bilinear pooling and attention schemes allow to
implement `a needle in a haystack' approaches, where each entry in the CNN
activations map is adaptively weighted. Thus, allowing to numerically
emphasize the contribution of the task-informative image locations, in
contrast to the visual clutter. For instance, consider the attention weights
visualization in Fig. \ref{fig:teaser}, where we highlight the position and
orientation of informative image locations. In this work, we utilize \textit{%
self-attention} where the inner products are computed between the different
entries of each particular activation map. This is implemented by an \textit{%
Transformer encoder}, in contrast to using a \textit{Transformer decoder }%
where\textit{\ }the inner products between the input data and an additional
set of vectors, denoted as \textit{queries}, are computed.

\section{Transformer Encoder for Pose Regression}

\label{sec:Localization regression}

In this work we propose to localize a camera by estimating $\mathbf{p} =<%
\mathbf{x},\mathbf{q}>$, where $\mathbf{x}\in \mathbb{R}^{3}$ is the
position of the camera in the world and $\mathbf{q}\in \mathbb{R}^{4}$ is
the quaternion encoding its 3D\ orientation. Thus, the problem of camera
pose regression is to jointly learn two regression tasks: regressing the
camera position $\mathbf{x}$ and regressing the camera orientation $\mathbf{q%
}$. We model each task as a sequence-to-one problem, where the input is a
sequential representation of a learned activation map and the output is an
estimate of the corresponding pose component (either $\mathbf{x}$ or $%
\mathbf{q}$). Following the success of Transformers in learning
sequence-to-one and sequence-to-sequence problems in NLP \cite%
{radford2019language} and more recently in Computer Vision \cite{DETR}, we
propose to use Transformer encoders for aggregating sequential representations
into a single latent vector. Specifically, we employ positional and orientational encoders that learn to emphasize the visual cues for position and orientation inference, respectively. An MLP can then regress the resulting encoder output, replacing the classifier head used in standard sequence-to-one
architectures \cite{bert}. We denote our model \textit{TransPoseNet} as it
shifts from the conventional pose regression paradigm to a Transformer-based
model. An overview of the proposed scheme is shown in Fig. \ref{fig:teaser},
where the image is initially processed by a CNN backbone to compute
activation maps that are analyzed by the dual encoders and corresponding
regression heads.

\subsection{Network Architecture}

\label{subsec:model_architecture}

TransPoseNet is composed of a shared convolutional backbone and two branches
for separately regressing the position and orientation of the camera. Each
branch consists of a separate Transformer encoder and an MLP head. Using
separate activation maps and Transformers allows us to attend to the
different features at different resolutions depending on the particular
learned task.

\textbf{Convolutional Backbone.} The activation maps of convolutional
backbones capture latent visual features at different resolutions. Given an
image $\mathbf{I}\in \mathbb{R}^{H\times W\times C}$, we sample the backbone
at two different resolutions and take an activation map per regression task:
$M_{\mathbf{x}}$ and $M_{\mathbf{q}}$ respectively.

\textbf{Sequential Representations of Activation Maps.} We follow \cite{DETR}
and convert an activation map $\mathbf{M}\in \mathbb{R}^{H_{m}\times
W_{m}\times C_{m}}$ to a sequence $\widehat{\mathbf{M}}\in \mathbb{R}^{{H_{m}%
}\cdot W_{m}\times C_{t}}$ using a $1\times1$ convolution (projecting to
dimension $C_{t}$) followed by flattening . We further append a learned
\textit{orientation/position token} $\mathbf{t}\in \mathbb{R}^{C_{t}}$ to
the sequence of map features, analogous to the \textit{class token} in
sequence-to-one NLP classification Transformer architectures \cite{bert}.
Hence, the input to the encoder is given by:%
\begin{equation}
E_{in}=[\mathbf{t},\widehat{\mathbf{M}}]\in \mathbb{R}^{({H_{m}}\cdot
W_{m}+1)\times C_{t}}.
\end{equation}

\textbf{Positional Encoding.} In order to preserve the spatial information
of each location in the map, and assign a particular positional encoding to
the token $\mathbf{t}$, the spatial positions are encoded using positional
encoding as in \cite{DETR}. We train two one-dimensional encodings for the $%
X $ and $Y$ axes separately, to reduce the number of learnt positional
parameters. Thus, for an activation map $\mathbf{M}$ we define the sets of
positional embedding vectors $\mathbf{E}_{x}\in \mathbb{R}^{\left(
W_{m}+1\right) \times C_{t}/2}$ and $\mathbf{E}_{y}\in \mathbb{R}^{\left(
H_{m}+1\right) \times C_{t}/2}$, such that a spatial position $\left(
i,j\right) ,$ $i\in 1..H_{m}$, $j\in 1..W_{m}$, is encoded by the
concatenating of the corresponding embedding vector:%
\begin{equation}
\mathbf{E}_{pos}^{i,j}=%
\begin{bmatrix}
\mathbf{E}_{x}^{j} \\
\mathbf{E}_{y}^{i}%
\end{bmatrix}%
\in \mathbb{R}^{C_{t}}.
\end{equation}

The pose token is assigned with position $(0,0)$ and the embedding vectors $(%
\mathbf{E}_{x}^{0},\mathbf{E}_{y}^{0})$, as these vectors will only be
associated with that token. Similar to the activation map $\mathbf{M}$, the
positional embeddings are reshaped as sequence $\widehat{\mathbf{E}}\in
\mathbb{R}^{({H_{m}}\cdot W_{m}+1)\times C_{t}}$ and passed to the encoder.

\textbf{Transformer Encoder.} We use a standard Transformer encoder
architecture \cite{bert} consisting of $n$ identical blocks. Each block
contains a self multi-head attention (MHA) module and an MLP with two layers
and gelu non-linearity. The input is passed through a LayerNorm \cite%
{AttentionIsAllYouNeed} before each module (MHA/MLP) and added back to the
output with residual connections \cite{AttentionIsAllYouNeed} and dropout. As in \cite{DETR} we add the position encoding to the input before each layer and normalize the output of the final layer with an additional LN pass.
The encoder's output $\mathbf{t}^{\prime }\in \mathbb{R}^{C_{t}}$ at the
position of the special token $t$, represents a global and context aware
summarization of the local features of the input activation map.

\textbf{MLP Head.} an MLP head with one hidden layer and gelu non-linearity
regresses the target vector ($\mathbf{x}$ or $\mathbf{q}$) from $\mathbf{%
t^{\prime}}$.

\subsection{Camera Pose Loss}

\label{subsec:loss}

Camera pose regressors are optimized to minimize the deviation between the
ground truth pose $\mathbf{p}_{0}=<\mathbf{x}_{0},\mathbf{q}_{0}>$ and the
predicted pose $\mathbf{p}=<\mathbf{x},\mathbf{q}>$. The position loss $L_{%
\mathbf{x}}$ and the orientation loss $L_{\mathbf{q}}$ are measured by the
Euclidean distance between the ground truth and the estimation:
\begin{equation}
L_{\mathbf{x}}=||\mathbf{x}_{0}-\mathbf{x}||_{2}  \label{equ:position loss}
\end{equation}%
\begin{equation}
L_{\mathbf{q}}=\left\Vert \mathbf{q_{0}}-\frac{\mathbf{q}}{||\mathbf{q}||}%
\right\Vert _{2}  \label{equ:orientation loss}
\end{equation}%
where $\mathbf{q}$ is normalized to unit vector in order to map to a valid
rotation matrix. In order to balance between the two losses, while
considering their scale differences, Kendall et al. suggested to learn their
relative weight by modelling the uncertainty of each task \cite%
{kendall2017geometric}:%
\begin{equation}
L_{\mathbf{p}}=L_{\mathbf{x}}\exp (-s_{\mathbf{x}})+s_{\mathbf{x}}+L_{%
\mathbf{q}}\exp (-s_{\mathbf{q}})+s_{\mathbf{q}},
\label{equ:learnable pose loss}
\end{equation}%
where $s_{\mathbf{x}}$ and $s_{\mathbf{q}}$ are learned parameters.
Recently, Shavit and Ferens showed the advantage of separately learning each
task on its own \cite{shavitferensirpnet}. Here, we follow a combined
approach, where we first train the entire model to minimize Eq. \ref%
{equ:learnable pose loss} and then fine-tune each MLP head solely with its
respective loss (Eqs. \ref{equ:position loss} and \ref{equ:orientation loss}%
).

\subsection{Implementation Details}

Our localization approach is implemented using EfficientNet-B0 \cite%
{pmlr-v97-tan19a} as our convolutional backbone, previously shown to focus
on informative visual features for image classification. The EfficientNet
backbone has several reduction levels, each corresponding to activation maps
with decreasing resolution and increasing depth. We use an open source
implementation of EfficientNet \cite{efficientnet-pytorch} and use the
activation maps from two endpoints (reduction levels): $m_{rdct4}\in \mathbb{%
R}^{14\times 14\times 112}$ and $m_{rdct3}\in \mathbb{R}^{28\times 28\times
40}$. We use $m_{rdct4}$ and $m_{rdct3}$ as inputs for the position and
orientation branches, respectively. We linearly project each activation map
to a common depth dimension $C_{t}=256$, and learn positional encoding of
the same depth. For the Transformer encoder we use a standard implementation
with six blocks and a dropout of $p=0.1$. Each block contains an MHA layer
with four heads and an MLP preserving the input dimension $C_{t}$. The
hidden layer of the MLP regressor heads expands the dimension from $256$ to $%
1024$ before regressing the output vector with a fully connected layer.

\section{Experimental Results}

\label{experiments}
\begin{table*}[tbh]
\caption{Comparative analysis of pose regressors on the Cambridge Landmarks
dataset (outdoor localization). We report the median position/orientation
error in meters/degrees for each method. Best results are highlighted in
bold.}
\label{tb:cambridge_res}\centering
\begin{tabular}{cccccc}
\hline
\textbf{Method} & \textbf{K. College} & \textbf{Old Hospital} & \textbf{Shop
Facade} & \textbf{St. Mary} &  \\ \hline
\multicolumn{1}{l}{IR Baseline \cite{sattler2019understanding}} & 1.48/4.45
& 2.68/4.63 & 0.90/4.32 & 1.62/6.06 &  \\
\multicolumn{1}{l}{PoseNet \cite{kendall2015posenet}} & 1.92/5.40 & 2.31/5.38
& 1.46/8.08 & 2.65/8.48 &  \\
\multicolumn{1}{l}{BayesianPN \cite{kendall2016modelling}} & 1.74/4.06 &
2.57/5.14 & 1.25/7.54 & 2.11/8.38 &  \\
\multicolumn{1}{l}{LSTM-PN \cite{walch2017image}} & 0.99/3.65 & 1.51/4.29 &
1.18/7.44 & 1.52/6.68 &  \\
\multicolumn{1}{l}{SVS-Pose \cite{naseer2017deep}} & 1.06/2.81 & 1.50/4.03 &
0.63/5.73 & 2.11/8.11 &  \\
\multicolumn{1}{l}{GPoseNet \cite{cai2019hybrid}} & 1.61/2.29 & 2.62/3.89 &
1.14/5.73 & 2.93/6.46 &  \\
\multicolumn{1}{l}{PoseNet-Learnable \cite{kendall2017geometric}} & 0.99/1.06
& 2.17/\textbf{2.94} & 1.05/3.97 & 1.49/3.43 &  \\
\multicolumn{1}{l}{GeoPoseNet \cite{kendall2017geometric}} & 0.88/\textbf{%
1.04} & 3.20/3.29 & 0.88/3.78 & 1.57/\textbf{3.32} &  \\
\multicolumn{1}{l}{MapNet \cite{brahmbhatt2018geometry}} & 1.07/1.89 &
1.94/3.91 & 1.49/4.22 & 2.00/4.53 &  \\
\multicolumn{1}{l}{IRPNet \cite{shavitferensirpnet}} & 1.18/2.19 & 1.87/3.38
& 0.72/3.47 & 1.87/4.94 &  \\
\multicolumn{1}{l}{\textbf{TransPoseNet (Ours)}} & \textbf{0.60}/2.43 &
\textbf{1.45}/3.08 & \textbf{0.55/3.49} & \textbf{1.09}/4.99 &  \\ \hline
\end{tabular}%
\end{table*}
Our proposed scheme focuses on improving the accuracy of \textit{absolute pose regression}. As such, we evaluate it on multiple
contemporary datasets used for benchmarking camera pose regressors, and
compare the results to recent state-of-the-art regression-based absolute
localization methods. Other classes of localization schemes,  described in section \ref{subsec:related pose estimation}, which utilize additional data at inference time (localization pipelines \cite%
{taira2018inloc,sarlin2019coarse,noh2017large,dusmanu2019d2} and relative
pose regression \cite{balntas2018relocnet,laskar2017camera,ding2019camnet}) or that are an order of magnitude slower (3D-based scene coordinate regression \cite{DSAC,DSAC++}) are not considered for this analysis.

\textbf{Datasets.} The Cambridge Landmarks \cite{kendall2015posenet} dataset
depicts an urban environment and represents outdoor localization tasks with
a spatial extent of $\sim900-5500m^{2}$. We use four of its six scenes for
comparative evaluation, as the other two scenes are not commonly
benchmarked. The 7Scenes \cite{glocker2013real} dataset includes seven
small-scale indoor scenes, depicting a spatial extent of $\sim1-10m^{2}$.
Both datasets present various localization challenges such as occlusions,
reflections, motion blur, lighting conditions, repetitive textures and
variations in view point and trajectory.

\textbf{Training.} We optimize our model using Adam, with $\beta _{1}=0.9$, $%
\beta _{2}=0.999$ and $\epsilon =10^{-10}$ and a batch size of $8$. At the
first step of the training we minimize the loss defined in Eq. \ref%
{equ:learnable pose loss}. Loss parameters are initialized as in \cite%
{valada2018deep}. We use a learning rate of $\lambda =10^{-4}$ and decrease
it by a factor of $10$ every $100$ ($200$) epochs for indoor (outdoor)
localization for up to $300$ ($600$) epochs. We apply a weight decay of $%
10^{-4}$ and train the encoders with a dropout $p=0.1$. At the second
training stage, we fine-tune each MLP head separately with its respective
loss (Eqs. \ref{equ:position loss} and \ref{equ:orientation loss}), while
freezing all other weights in the network. This allows us to achieve better
performance without needing to trade-off between the two objectives. Since
the position is dominant in localizing large scale scenes, we fine-tune the
orientation head with a latent position prior from the position Transformer
encoder (changing the orientation MLP to take a concatenated input). For
indoor localization, we train both heads independently and use a higher
learning rate and a higher weight decay ($10^{-3}$ and $10^{-2}$,
respectively). In order to allow our model to better generalize, we follow
the augmentation procedure used by \cite{kendall2015posenet}. At train time
we rescale the image so that its smaller edge is resized to $256$ pixels and
take a random $224\times 224$ crop. We then randomly jitter the brightness,
contrast and saturation. At test time we take the center crop after
rescaling with no further augmentations. All models were trained on a single
NVIDIA Tesla V100 GPU using the PyTorch framework \cite{paszke2019pytorch}.

\subsection{Comparative Analysis of Camera Pose Regressors}

\label{subsec:Comparative}
\begin{table}[tbh]
\caption{Localization results for the Cambridge Landmarks dataset. We report
the average position/orientation errors in meters/degrees and the respective
rankings. Best results are highlighted in bold. The final ranking is set
according to the average rank.}
\label{tb:cambridge_rank}\centering
\begin{tabular}{cccc}
\hline
\textbf{Method} &
\begin{tabular}{@{}c}
\textbf{Average} \\
\textbf{[m/deg]}%
\end{tabular}
& \textbf{Ranks} &
\begin{tabular}{@{}c}
\textbf{Final} \\
\textbf{Rank}%
\end{tabular}
\\ \hline
\multicolumn{1}{l}{PoseNet \cite{kendall2015posenet}} & 2.09/6.84 & 10/10 &
10 \\
\multicolumn{1}{l}{BayesianPN \cite{kendall2016modelling}} & 1.92/6.28 & 8/9
& 9 \\
\multicolumn{1}{l}{LSTM-PN \cite{walch2017image}} & 1.30/5.52 & 2/8 & 5 \\
\multicolumn{1}{l}{SVS-Pose \cite{naseer2017deep}} & 1.33/5.17 & 3/7 & 5 \\
\multicolumn{1}{l}{GPoseNet \cite{cai2019hybrid}} & 2.08/4.59 & 9/6 & 8 \\
\multicolumn{1}{l}{PoseNet-Learnable \cite{kendall2017geometric}} & 1.43/%
\textbf{2.85} & 5/\textbf{1} & 2 \\
\multicolumn{1}{l}{GeoPoseNet \cite{kendall2017geometric}} & 1.63/2.86 & 6/2
& 3 \\
\multicolumn{1}{l}{MapNet \cite{brahmbhatt2018geometry}} & 1.63/3.64 & 6/5 &
7 \\
\multicolumn{1}{l}{IRRNet \cite{shavitferensirpnet}} & 1.42/3.45 & 4/4 & 3
\\
\multicolumn{1}{l}{\textbf{TransPoseNet (Ours)}} & \textbf{0.91}/3.47 &
\textbf{1}/3 & \textbf{1} \\ \hline
\end{tabular}%
\end{table}
\begin{table*}[tbh]
\caption{Comparative analysis of pose regressors on the 7Scenes dataset
(indoor localization). We report the median position/orientation error in
meters/degrees for each method. Best results are highlighted in bold.}
\label{tb:7scenes_res}\centering
\begin{tabular}{cccccccc}
\hline
\textbf{Method} & \textbf{Chess} & \textbf{Fire} & \textbf{Heads} & \textbf{%
Office} & \textbf{Pumpkin} & \textbf{Kitchen} & \textbf{Stairs} \\ \hline
\multicolumn{1}{l}{IR Baseline \cite{sattler2019understanding}} & 0.18/10.0
& 0.33/12.4 & 0.15/14.3 & 0.25/10.1 & 0.26/9.42 & 0.27/11.1 & \textbf{0.24}%
/14.7 \\
\multicolumn{1}{l}{PoseNet \cite{kendall2015posenet}} & 0.32/8.12 & 0.47/14.4
& 0.29/12.0 & 0.48/7.68 & 0.47/8.42 & 0.59/8.64 & 0.47/13.8 \\
\multicolumn{1}{l}{BayesianPN \cite{kendall2016modelling}} & 0.37/7.24 &
0.43/13.7 & 0.31/12.0 & 0.48/8.04 & 0.61/7.08 & 0.58/7.54 & 0.48/13.1 \\
\multicolumn{1}{l}{LSTM-PN \cite{walch2017image}} & 0.24/5.77 & 0.34/11.9 &
0.21/13.7 & 0.30/8.08 & 0.33/7.0 & 0.37/8.83 & 0.40/13.7 \\
\multicolumn{1}{l}{GPoseNet \cite{cai2019hybrid}} & 0.20/7.11 & 0.38/12.3 &
0.21/13.8 & 0.28/8.83 & 0.37/6.94 & 0.35/8.15 & 0.37/12.5 \\
\multicolumn{1}{l}{PoseNet-Learnable \cite{kendall2017geometric}} & 0.14/4.50
& 0.27/11.8 & 0.18/12.1 & 0.20/5.77 & 0.25/4.82 & 0.24/5.52 & 0.37/10.6 \\
\multicolumn{1}{l}{GeoPoseNet \cite{kendall2017geometric}} & 0.13/4.48 &
0.27/11.3 & 0.17/13.0 & 0.19/5.55 & 0.26/4.75 & 0.23/5.35 & 0.35/12.4 \\
\multicolumn{1}{l}{MapNet \cite{brahmbhatt2018geometry}} & \textbf{0.08/3.25}
& 0.27/11.7 & 0.18/13.3 & \textbf{0.17/5.15} & 0.22/\textbf{4.02} & 0.23/%
\textbf{4.93} & 0.30/12.1 \\
\multicolumn{1}{l}{IRPNet \cite{shavitferensirpnet}} & 0.13/5.64 & 0.25/%
\textbf{9.67} & 0.15/13.1 & 0.24/6.33 & 0.22/5.78 & 0.30/7.29 & 0.34/11.6 \\
\multicolumn{1}{l}{AttLoc \cite{wang2020atloc}} & 0.10/4.07 & 0.25/11.4 &
0.16/\textbf{11.8} & \textbf{0.17}/5.34 & 0.21/4.37 & 0.23/5.42 & 0.26/10.5
\\
\multicolumn{1}{l}{\textbf{TransPoseNet (Ours)}} & \textbf{0.08}/5.68 &
\textbf{0.24}/10.6 & \textbf{0.13}/12.7 & \textbf{0.17}/6.34 & \textbf{0.17}%
/5.6 & \textbf{0.19}/6.75 & 0.30/\textbf{7.02} \\ \hline
\end{tabular}%
\end{table*}
\begin{table}[tbh]
\caption{Localization results for the 7Scenes dataset. We report the average
position/orientation errors in meters/degrees and the respective rankings.
Best results are highlighted in bold. The final ranking is set according to
the average rank.}
\label{tb:7scenes_rank}\centering
\begin{tabular}{cccc}
\hline
\textbf{Method} &
\begin{tabular}{@{}c}
\textbf{Average} \\
\textbf{[m/deg]}%
\end{tabular}
& \textbf{Ranks} &
\begin{tabular}{@{}c}
\textbf{Final} \\
\textbf{Rank}%
\end{tabular}
\\ \hline
\multicolumn{1}{l}{PoseNet \cite{kendall2015posenet}} & 0.44/10.4 & $9/10$ &
$10$ \\
\multicolumn{1}{l}{BayesianPN \cite{kendall2016modelling}} & 0.47/9.81 & $%
10/7$ & $9$ \\
\multicolumn{1}{l}{LSTM-PN \cite{walch2017image}} & 0.31/9.86 & $7/8$ & $7$
\\
\multicolumn{1}{l}{GPoseNet \cite{cai2019hybrid}} & 0.31/9.95 & $7/9$ & $8$
\\
\multicolumn{1}{l}{PoseNet-Learnable \cite{kendall2017geometric}} & 0.24/7.87
& $6/4$ & $5$ \\
\multicolumn{1}{l}{GeoPoseNet \cite{kendall2017geometric}} & 0.23/8.12 & $%
4/5 $ & $4$ \\
\multicolumn{1}{l}{MapNet \cite{brahmbhatt2018geometry}} & 0.21/7.78 & $3/2$
& $3$ \\
\multicolumn{1}{l}{IRPNet \cite{shavitferensirpnet}} & 0.23/8.49 & $4/6$ & $%
5 $ \\
\multicolumn{1}{l}{AttLoc \cite{wang2020atloc}} & 0.20/\textbf{7.56} & $2/$%
\textbf{1} & \textbf{1} \\
\multicolumn{1}{l}{\textbf{TransPoseNet (Ours)}} & \textbf{0.18}/7.78 &
\textbf{1}/$2$ & \textbf{1} \\ \hline
\end{tabular}%
\end{table}
\begin{figure*}[tbh]
\centering
\subfigure[]{\includegraphics[width=0.48%
\linewidth]{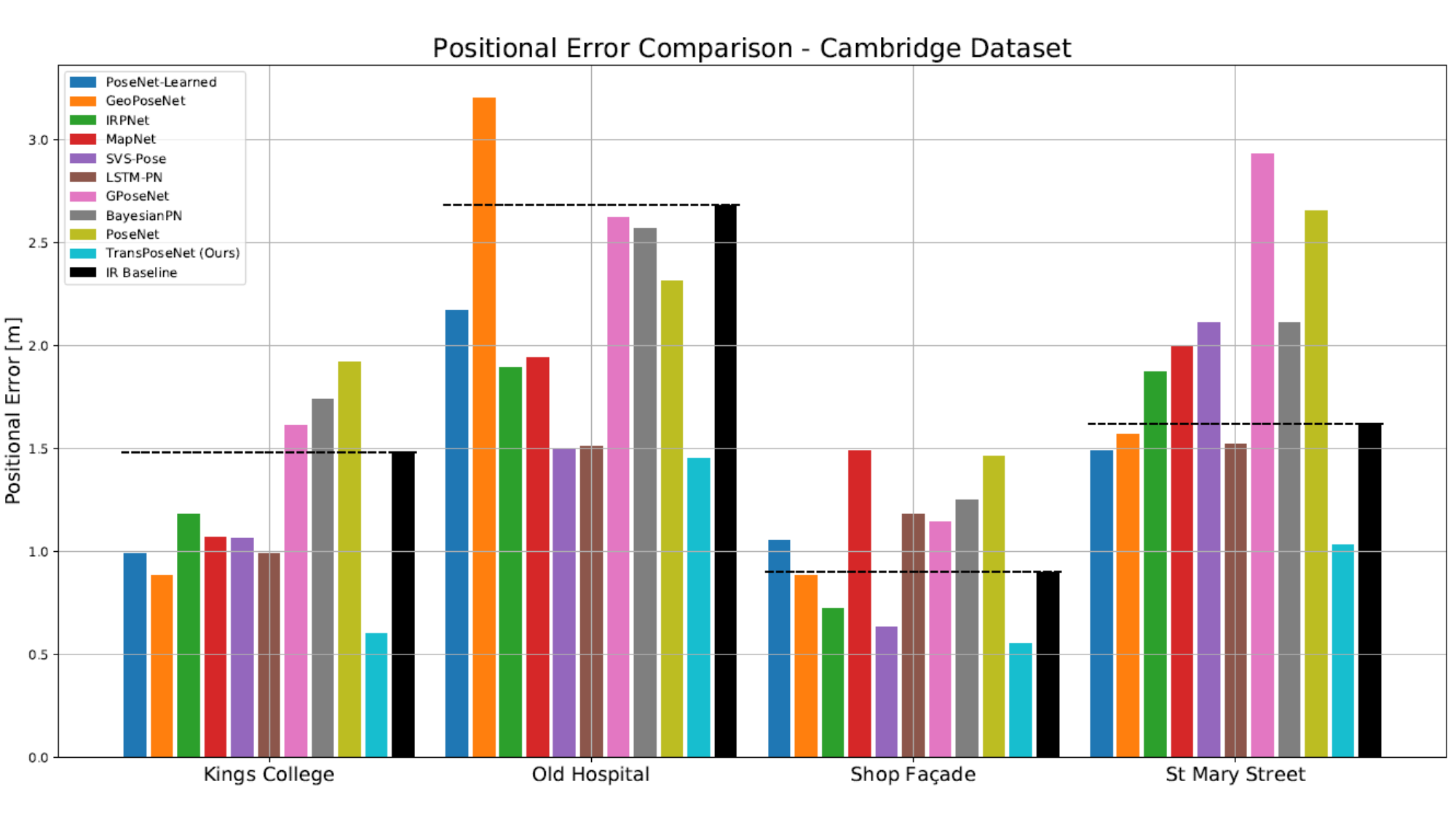}} \subfigure[]{%
\includegraphics[width=0.48%
\linewidth]{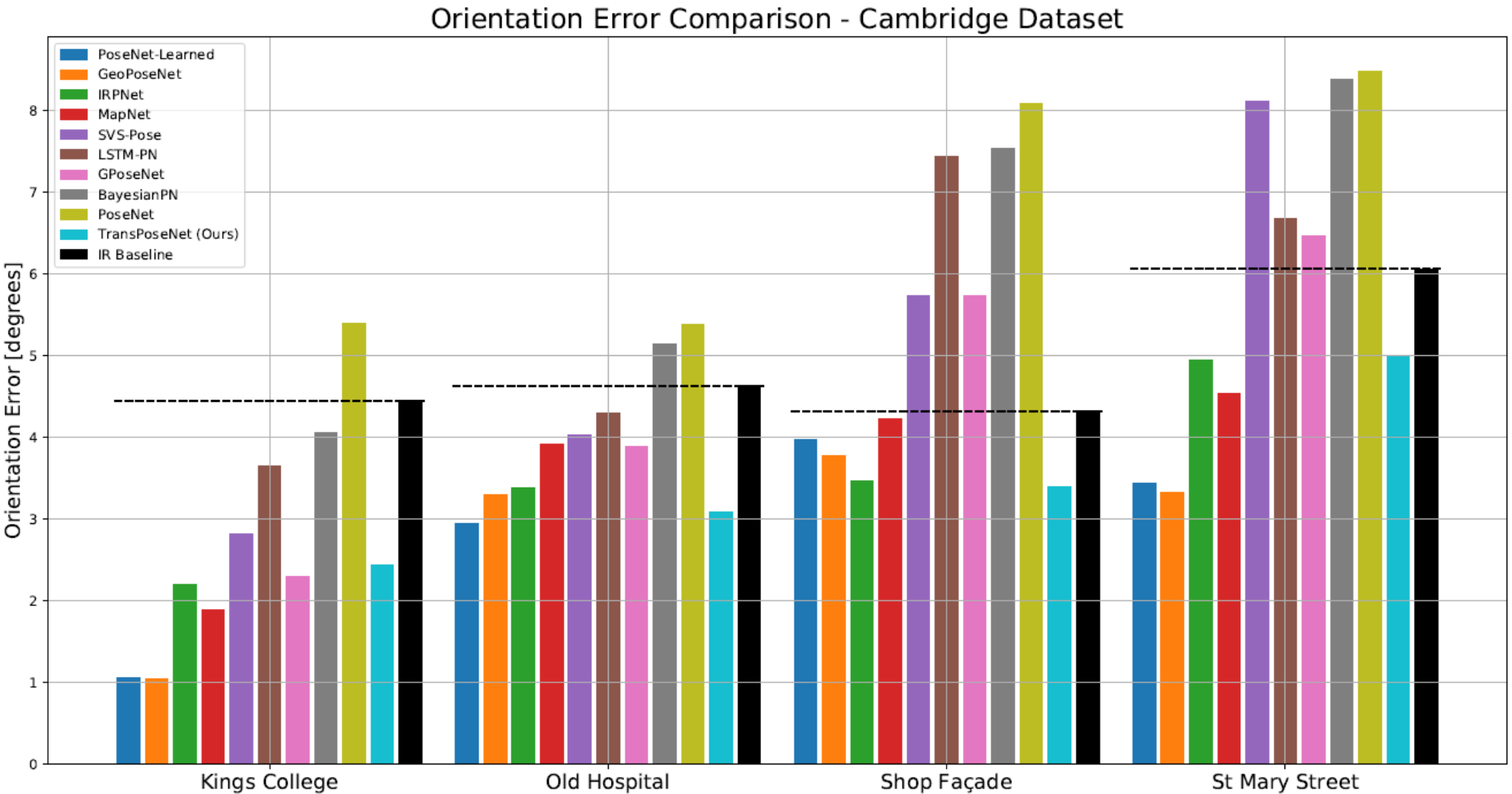}}
\caption{Visual comparison of position (a) and orientation (b) errors of
different pose regressors, with respect to the IR baseline reported by
\protect\cite{sattler2019understanding}. The results are reported for the
Cambridge Landmarks dataset.}
\label{fig_ir_baseline_cambridge}
\end{figure*}

\textbf{Outdoor Localization.} We report the median position and orientation
errors on four outdoor scenes from the Cambridge Landmarks dataset. Table %
\ref{tb:cambridge_res} shows the localization results of our method
(TransPoseNet) compared to all other pose regressors reporting on this
benchmark using a single input image. We also include the results of an IR
baseline recently proposed as a reference for evaluating the limitations of
absolute pose regression methods \cite{sattler2019understanding}.
TransPoseNet yields the lowest position error across all scenes and provides
the lowest orientation error when evaluated on the Shop Facade scene. It
also achieves the highest number of \textit{state-of-the-art} results across
the dataset. When summarizing the results (Table \ref{tb:cambridge_rank}),
TransPoseNet is the only regressor achieving sub-meter accuracy. It ranks
first in terms of average position error and third in orientation. In
addition, when considering the total rank (average of position and
orientation ranks), TransPoseNet again ranks first.

We also consider an IR
baseline, recently shown to present a lower error bound on the performance
of current pose regressors. Figures \ref{fig_ir_baseline_cambridge}a and \ref%
{fig_ir_baseline_cambridge}b compare the position and orientation errors of
different pose regressors with respect to the IR baseline results (indicated
in black). TransPoseNet is the only pose regressor whose localization error
is consistently below the IR error bar, across both position and
orientation. To the best of our knowledge, TransPoseNet is the only pose
regressor to-date to achieve this result.

\textbf{Indoor Localization.} Table \ref{tb:7scenes_res} shows the results
obtained when evaluating TransPoseNet on the 7Scenes dataset. As in outdoor
localization we compare its performance to all pose regressors reporting on
this benchmark with a single image input. TransPoseNet achieves the best
positional error in six out of the seven scenes and the highest number of
\textit{state-of-the-art} results overall. Similar to outdoor localization,
TransPoseNet ranks favorably in terms of position and orientation errors,
achieving first and second place respectively and ranking first in total
(Table \ref{tb:7scenes_rank}). Another attention-based method \cite%
{wang2020atloc} also achieves the first place in total ranking, but with
fewer \textit{state-of-the-art} results (two versus seven achieved with
TransPoseNet). When comparing the performance to the IR baseline (Table \ref%
{tb:7scenes_res}), TransPoseNet has 14 out of 15 results under the
bar, more than any other pose regressor to-date.

\subsection{Paying Attention to Local Features}

\begin{figure}[bth]
\centering
\includegraphics[width=\linewidth]{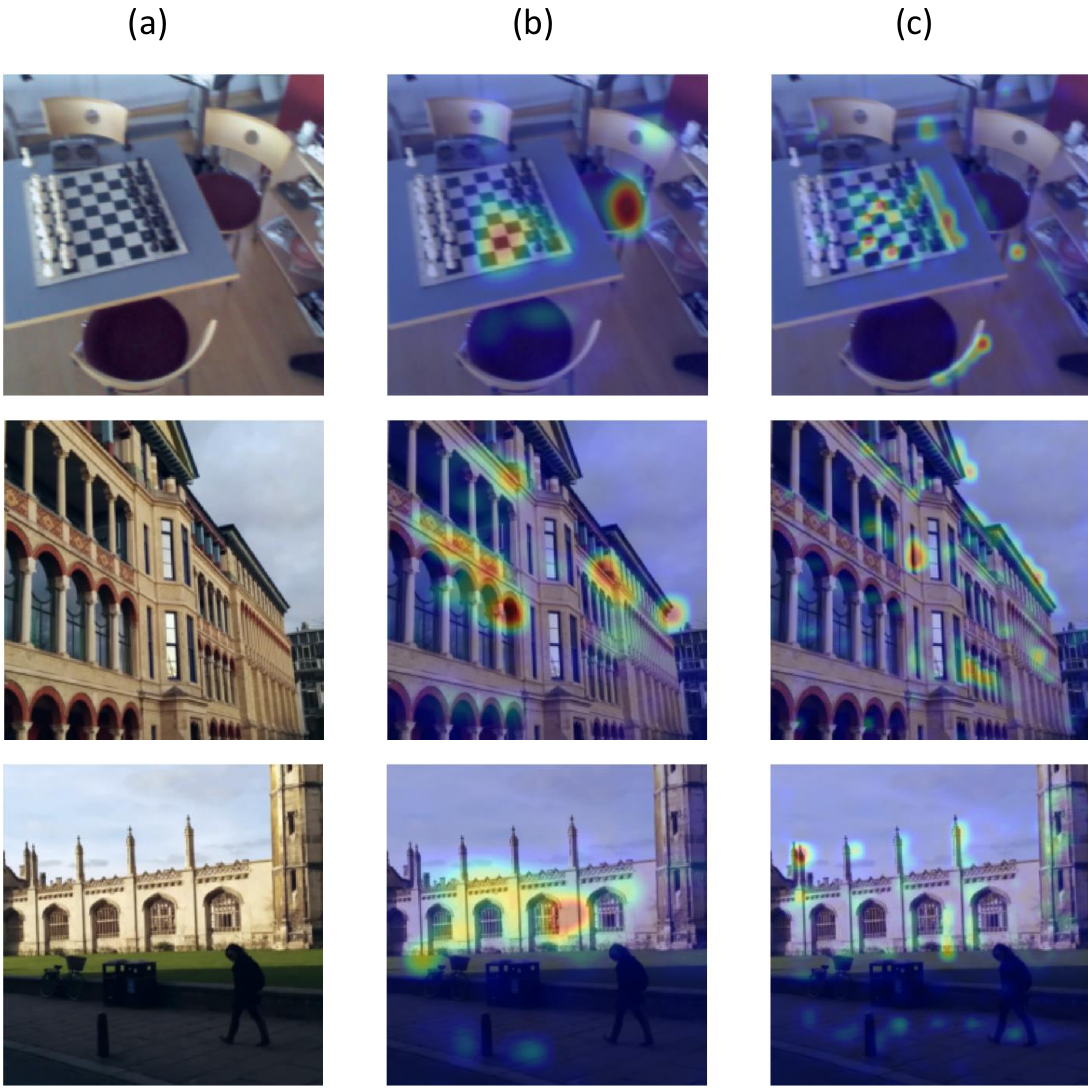}
\caption{Heatmap visualization of attention weights from the position and
orientation encoders (column (b) and (c)). Each row shows the query image
(column (a)) along with the respecting attentional heatmaps.}
\label{fig3_attention_maps}
\end{figure}

A key element in our proposed method is the use of one encoder per task
(position and orientation) for processing activation maps. The motivation
behind this design is to allow each encoder to attend to different features
in the image. We assume that the network will focus on features such as
corners and blobs to estimate the relative translation, and on edges and
lines for estimating the relative rotation. Figure \ref{fig3_attention_maps}
shows attention heatmaps for three query images taken from different scenes
in the Cambridge Landmarks and the 7Scenes datasets. Each row depicts the
query image and the upsampled attention weights taken from the last layers
of the position and orientation encoders (second and third column,
respectively). The position encoder tends to assign higher weights to
features that are intuitively more distinctive for coarser localization and
position estimation. In contrast, the orientation encoder focuses on lines
and long edges that are informative for estimating the rotational motion.
For example, consider the triplet in the third row, the position encoder
focuses on the windows at the image's center of mass, while the orientation
encoder attends to the elongated spires. The observed differences between
the two encoders reinforces our dual-head model architecture. The relative
motion emerging from attention-driven local features in the image, with
respect to a memorized map, can be summarized with context-aware global
absolute image descriptors used for regressing the position and orientation.

\subsection{Ablation Study}

We study the importance and impact of two aspects of our model design: the
attention mechanism and the resolution and depth of the activation maps.  In each evaluation, we started from the network used in Section \ref{subsec:Comparative} and modified a \textit{%
single} algorithmic component or hyper-parameter. All ablations were conducted
using the Shop Facade scene.
\begin{table}[tbh]
\caption{Ablation study of CNN backbones evaluated on the Shop Facade scene.
We report the median position and orientation errors. Both EfficientNet-based CNNs outperform the Resnet50 CNN. Using the EfficientNetB1 backbone degrades the accuracy due to overfitting.}
\label{table:ablation-cnn}\centering%
\begin{tabular}{lll}
\hline
\textbf{Backbone} & \textbf{Position } & \textbf{Orientation} \\
\multicolumn{1}{c}{} & \multicolumn{1}{c}{\textbf{[meters]}} & \textbf{%
[degrees]} \\ \hline
\multicolumn{1}{c}{Resnet50} & \multicolumn{1}{c}{0.81} & \multicolumn{1}{c}{
4.76} \\
\multicolumn{1}{c}{\textbf{EfficientNetB0}} & \multicolumn{1}{c}{\textbf{0.65%
}} & \multicolumn{1}{c}{\textbf{3.63}} \\
\multicolumn{1}{c}{EfficientNetB1} & \multicolumn{1}{c}{0.69} &
\multicolumn{1}{c}{4.03} \\ \hline
\end{tabular}%
\end{table}

\textbf{Backbone.} For studying the choice of backbone, we considered the
Resnet50, EfficientNetB0 and EfficientNetB1 CNNs, where both EfficientNet
variants use significantly less parameters than Resnet50. As shown in Table \ref{table:ablation-cnn}, both EfficientNet
variants significantly outperforms Resnet50. It is unclear from our
simulations, whether this is due to less overfitting or to their improved
architecture. The smaller EfficientNetB0 backbone provides the best results.

\textbf{The Role of Attention and Transformers.} To study the parameters of
the Transformer encoders, we trained our scheme with a varying number of
Transformer encoder layers. On the hand, the more layers are used, the better the learning capacity. On the other hand, an over-parameterized Transformer encoder might lead to overfitting. The results reported in Table \ref%
{table:ablation-encoder-layers} show a `sweet spot' when using six
layers. We also varied the Transformer encoders' inner embedding dimension to verify
our choice. The results in Table \ref{table:ablation-encoder-dims} show
that, similar to varying the number of layers, there exists a 'sweet spot' when using 256 for the encoder latent dimension.
\begin{table}[tbh]
\caption{Ablations of the Transformer Encoder configuration evaluated on the Shop Facade scene.
We report the median position and orientation errors. Using six layers optimizes the accuracy and avoids overfitting.}
\label{table:ablation-encoder-layers}\centering%
\begin{tabular}{lll}
\hline
\textbf{\#Encoder Layers} & \textbf{Position } & \textbf{Orientation} \\
\multicolumn{1}{c}{} & \multicolumn{1}{c}{\textbf{[meters]}} & \textbf{%
[degrees]} \\ \hline
\multicolumn{1}{c}{2} & \multicolumn{1}{c}{0.92} & \multicolumn{1}{c}{4.61}
\\
\multicolumn{1}{c}{4} & \multicolumn{1}{c}{0.93} & \multicolumn{1}{c}{4.2}
\\
\multicolumn{1}{c}{\textbf{6}} & \multicolumn{1}{c}{\textbf{0.65}} &
\multicolumn{1}{c}{\textbf{3.63}} \\
\multicolumn{1}{c}{8} & \multicolumn{1}{c}{0.89} & \multicolumn{1}{c}{3.75}
\\ \hline
\end{tabular}%
\end{table}
\begin{table}[tbh]
\caption{Ablations of model's embedding dimensionality, evaluated on the Shop Facade scene.
We report the median position and orientation errors.}
\label{table:ablation-encoder-dims}\centering
\begin{tabular}{lll}
\hline
\textbf{Encoder Dimension} & \textbf{Position } & \textbf{Orientation} \\
\multicolumn{1}{c}{} & \multicolumn{1}{c}{\textbf{[meters]}} & \textbf{%
[degrees]} \\ \hline
\multicolumn{1}{c}{64} & \multicolumn{1}{c}{0.79} & \multicolumn{1}{c}{4.35}
\\
\multicolumn{1}{c}{128} & \multicolumn{1}{c}{0.81} & \multicolumn{1}{c}{3.99}
\\
\multicolumn{1}{c}{\textbf{256}} & \multicolumn{1}{c}{\textbf{0.65}} &
\multicolumn{1}{c}{\textbf{3.63}} \\
\multicolumn{1}{c}{512} & \multicolumn{1}{c}{0.68} & \multicolumn{1}{c}{4.18}
\\ \hline
\end{tabular}%
\end{table}

\textbf{Localizing with Coarse and Fine Activation Maps.} Activation maps
from different endpoints of a CNN backbone capture different features in the
input image. Deeper reduction levels cover growing areas in the figure with
an increased depth. As orientation and position estimation may require
different feature complexity, we train our model with different
combinations of $m_{rdct3}\in \mathbb{R}^{28\times 28\times 40}$ and $%
m_{rdct4}\in \mathbb{R}^{14\times 14\times 112}$ from the EfficientNet
backbone, as inputs for the position and orientation encoders (without
fine-tuning). Table \ref{tb:reduct_comparative_analysis} shows the results
of four combinations evaluated on the Shop Facade scene from the Cambridge
Landmarks dataset. We either provide $m_{rdct4}$/$m_{rdct3}$ to both encoder
heads or use a combination (providing $m_{rdct3}$ to one encoder and $%
m_{rdct4}$ to the other). Assigning the position encoder with coarser
activation maps yields better accuracy. An opposite trend is observed for
activation maps used by the orientation encoder.
\begin{table}[tbh]
\caption{Ablations of activation maps evaluated on the Shop Facade scene. We
provide different combinations of $m_{rdct3}\in \mathbb{R}^{28\times
28\times 40}$ and $m_{rdct4}\in \mathbb{R}^{14\times 14\times 112}$ to the
encoder heads and report the median position and orientation errors.}
\label{tb:reduct_comparative_analysis}\centering
\begin{tabular}{ccc}
\hline
\begin{tabular}{@{}c}
\textbf{Reduction} \\
\textbf{Position/Orientation}%
\end{tabular}
&
\begin{tabular}{@{}c}
\textbf{Position} \\
\textbf{[meters]}%
\end{tabular}
&
\begin{tabular}{@{}c}
\textbf{Orientation} \\
\textbf{[degrees]}%
\end{tabular}
\\ \hline
\multicolumn{1}{l}{$m_{rdct3}$/$m_{rdct3}$} & 1.08 & 3.60 \\
\multicolumn{1}{l}{$m_{rdct3}$/$m_{rdct4}$} & 0.99 & 4.32 \\
\multicolumn{1}{l}{$\mathbf{m}_{\mathbf{rdct4}}$\textbf{/}$\mathbf{m}_{%
\mathbf{rdct3}}$} & \textbf{0.65} & \textbf{3.63} \\
\multicolumn{1}{l}{$m_{rdct4}$/$m_{rdct4}$} & 0.81 & 4.05 \\ \hline
\end{tabular}%
\end{table}
\section{Conclusion}

We presented a novel paradigm for camera pose regression which uses
sequential representations of multi-scale activation maps instead of a
single global descriptor from a CNN backbone. By applying dual multi-head
attention encoders, our model is able to focus on different local features
depending on the learned task. Our method ranks first on both outdoor and
indoor benchmarks and achieves a more consistent performance with respect to
recent baselines.

{\small
\bibliographystyle{ieee_fullname}
\bibliography{TransPoseNet_arxiv}
}

\end{document}